\documentclass[review]{elsarticle}

\usepackage{lineno,hyperref}
\usepackage{subcaption}
\usepackage{graphicx}
\usepackage{caption}
\usepackage{multirow}
\usepackage{adjustbox}
\usepackage{tabularx}
\usepackage{url}
\usepackage{amsmath}
\usepackage{amssymb}
\usepackage[dvipsnames]{xcolor}
\usepackage[ruled]{algorithm2e}
\usepackage{algcompatible}
\usepackage{scrextend}
\usepackage{todonotes}
\usepackage{lipsum}
\usepackage{amsthm}
\usepackage{enumitem}
\usepackage{wrapfig}
\usepackage{multicol}
\usepackage{amsthm}
\usepackage{sidecap}
\usepackage{soul}
\usepackage[a4paper,margin=1in,footskip=0.25in]{geometry}

\def\bs{\boldsymbol}
\algdef{SE}[SUBALG]{Indent}{EndIndent}{}{\algorithmicend\ }%
\algtext*{Indent}
\algtext*{EndIndent}

\newtheorem{definition}{Definition}[section]

\newcolumntype{P}[1]{>{\centering\arraybackslash}p{#1}}

\journal{Journal of Pattern Recognition}

\begin{document}

\begin{frontmatter}

\title{Reconstruction of Fragmented Trajectories of Collective Motion using Hadamard Deep Autoencoders}

\author[tamucc]{Kelum~Gajamannage\corref{mycorrespondingauthor}}
\cortext[mycorrespondingauthor]{Corresponding author}
\address[tamucc]{Department of Mathematics and Statistics, Texas A\&M University--Corpus Christi, Corpus Christi, TX--78412, USA.}
\ead{kelum.gajamannage@tamucc.edu}

\author[tamucc]{Yonggi~Park}
\ead{yonggi.park@tamucc.edu}

\author[wpi]{Randy~Paffenroth}
\address[wpi]{Department of Mathematical Sciences, Department of Computer Science, Data Science Program, Worcester Polytechnic Institute, Worcester, MA--01609, USA.}
\ead{rcpaffenroth@wpi.edu}

\author[csu]{Anura~P.~Jayasumana}
\address[csu]{Department of Electrical and Computer Engineering, Colorado State University, Fort Collins, CO--80525, USA.}
\ead{anura.jayasumana@colostate.edu}

%
%
%

\begin{abstract}
Learning dynamics of collectively moving agents such as fish or humans is an active field in research. Due to natural phenomena such as occlusion and change of illumination, the multi-object methods tracking such dynamics might lose track of the agents where that might result fragmentation in the constructed trajectories. Here, we present an extended deep autoencoder (DA) that we train only on fully observed segments of the trajectories by defining its loss function as the Hadamard product of a binary indicator matrix with the absolute difference between the outputs and the labels. The trajectories of the agents practicing collective motion is low-rank due to mutual interactions and dependencies between the agents that we utilize as the underlying pattern that our Hadamard deep autoencoder (HDA) codes during its training. The performance of our HDA is compared with that of a low-rank matrix completion scheme in the context of fragmented trajectory reconstruction.
\end{abstract}


\end{frontmatter}

\section{Introduction}
\label{sec:introduction}
{W}ith the advancements in both the video recording technology and the multi-object tracking (MOT) methods, dynamics of animal groups in wild \cite{Carter2013}, pedestrians on streets \cite{Pellegrini09}, vehicles on roads \cite{Betke2000}, and sport players on courts \cite{lu13} are studied using agents' individual trajectories. Utilization of trajectories are favorable for such studies rather than that of raw-videos since trajectories facilitate learning of dynamics of not only the entire group, but also that of either each individual agent or their clusters \cite{Gajamannage2015}.  MOT is an important computer vision subject that has received an increasing attention due to its broad applicability in both academic and commercial sectors  \cite{wenhan2021}.  In general, MOT methods undergo three steps: locating multiple objects, maintaining their identities, and constructing their individual trajectories \cite{Gajamannage2015}. However, due to natural phenomena such as occlusion, change of  illumination, and  similar appearance \cite{Gajamannage2019Reconstruction}, the trajectories produced by some MOT methods are severely fragmented with missing segments. 

MOT methods maintain a unique identity for each object and then map the objects having the same identity between time-steps to get full trajectories \cite{wenhan2021}. However, due to the aforesaid natural phenomena, tracking methods might ill-perform by either switching the identities or fragmenting the trajectories. Widely used remedies to treat such issues are namely part-to-whole, hypothesize-and-test, and buffer-and-recover \cite{wenhan2021}. Part-to-whole is based on the assumption that a part of the object is still visible when an occlusion happens; thus, this strategy observes and utilizes the visible part to infer the state of the whole object. Hypothesize-and-test guesses possible connections to treat the fragmentation and tests the validity of them based on the available observations. Buffer-and-recover stores observations and remembers the states of objects before an occlusion happens where the objects’ states are recovered based on the buffered observations. However, the above conventional remedies for treating these tracking issues are sometimes computationally expensive or/and less precise. 

We utilize a generalized version of deep autoencoders (DAs) to reconstruct fragmented trajectories. A deep autoencoder \cite{Liou2014}, is a type of artificial neural network (ANN) that has two main parts: an encoder that maps input data to a compressed version called \emph{code}, and a decoder that maps this code to the output \cite{cheng2018deep}. DAs have several intermediate layers and can be customized to data of interest by adjusting the features such as number of layers, layer size, code size, etc. Although the conventional DAs are not tailored to work with missing entries of input data, Hadamard treatment allows training an ANN only at the observed portion of the data. Especially, we reformulate the conventional loss-function of DAs that quantifies the reconstruction error with a binary indicator matrix consisting of ones and zeros corresponding to observed and unobserved entries, respectively, of the trajectories. The new loss function is defined as entry-wise product of DA's conventional loss function with the indicator matrix. Ref.~\cite{karkare2021blind} presents a robust DA by combining the conventional DA onto the robust principal component analysis scheme in \cite{candes2011robust} where this robust DA ensembles both the low-rank and the sparse components of the data matrix. This Hadamard treatment ignores missing entries from the optimization when the ANN is trained even though the input data consists missing entries. Since the Hadamard product between the loss-function of the DA and the indicator matrix is the essential step of this trajectory reconstruction method, we call it as Hadamard deep autoencoder and abbreviate as HDA. 

The key motivation of employing DA-based reconstruction method is that the DAs generate a low-dimensional nonlinear representation of the low-rank collective motion data after filtering out coarse features including noise. The interactions between the agents of a system can be simple, complex, or somewhere in between, and can exist among neighbors in space  or in time \cite{gajamannage2016}. Trajectories of the agents in collective motion can be represented as a high-dimensional data-cloud in Euclidean space where each point in this data-cloud is called a \emph{configuration vector} that is defined as the vector representing the positions of all the agents of the group at a given time-step \cite{SGE}. As the agents practicing collective motion are locally dependent upon each other, this high-dimensional data-cloud can often be projected onto a low-rank linear subspace called a manifold that describes the underlying dynamics of the collective motion. Rank of a dataset can be approximated either by a linear approach or by a nonlinear approach; thus, to distinguish those two versions, we name them as linear rank and nonlinear rank. The nonlinear rank is empirically quantified as the number of singular values (SVs) of the Gramian matrix of the geodesic distance matrix. The geodesic distance matrix is computed in two steps: first, we create a graph structure, based upon configuration vectors where each configuration vector represents the coordinates of the agents at a given time-step. We chose a fix number of nearest configurations to each configuration using Euclidean distance \cite{agarwal1999geometric}, and then transform the nearest neighbors into a graph structure by treating configurations as graph nodes and connecting each pair of nearest neighbors by an edge having the weight equal to the Euclidean distance between them. Second, we estimate the geodesic distance between the nodes as their \emph{shortest path distance in the graph} and make the geodesic distance matrix \cite{SGE}. A noteworthy fact of this approach is that the degrees of freedom of the multi-object system equivalent to its nonlinear low-rank \cite{Gajamannage2019Reconstruction}. While high coordination among the agents guarantees low degrees of freedom for the system, low coordination assures the opposite. The relation of the nonlinear low-rankness to the HDA is that the ``code'' of HDA is equivalent to this low-rank subspace; thus, we conjecture HDA to exhibit significant performance in reconstructing fragmented trajectories of collective motion.  

\begin{figure}[tp]
\centering
\includegraphics[width=.8\textwidth]{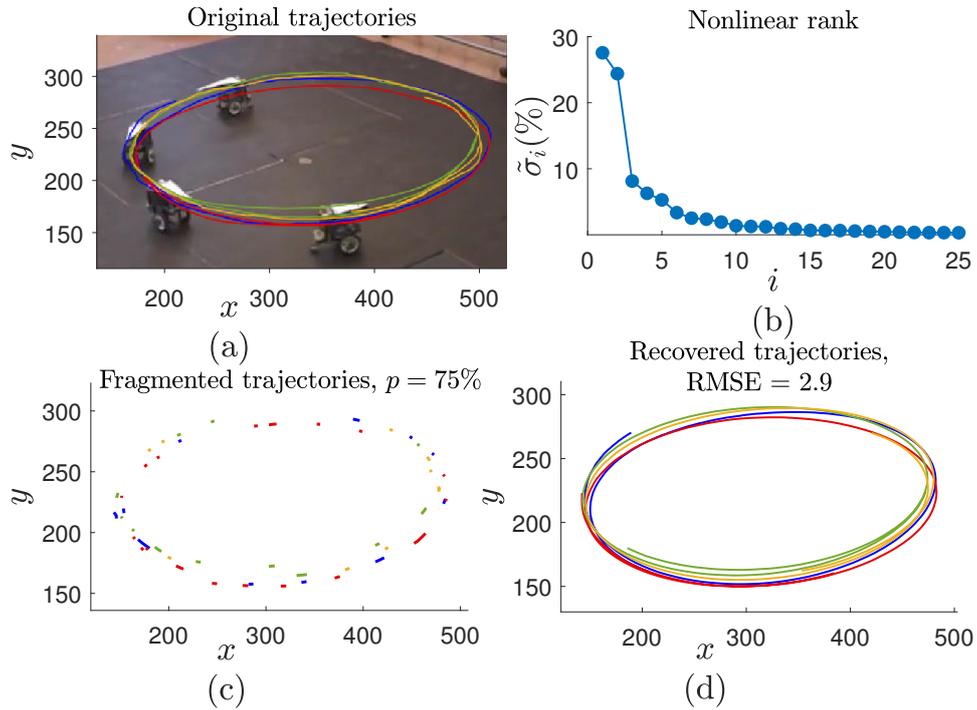}
\caption{Reconstruction of fragmented trajectories using Hadamard deep autoencoder with the aid of low-rankness of collective motion. (a) Trajectories of four mingling robots, are tracked using Kanade–Lucas–Tomasi tracker, that we show in red, blue, yellow, and green. (b) SV percentages of the Gramian matrix of the trajectory data, where $\tilde{\sigma}_i$ represents the percentage of the $i$-th SV, shows that only two SVs are prominent with approximately $50\%$ the total variance while the first 20 SVs contributes almost 100\% of the variance. To mimic the missing trajectory segments due to natural phenomena, (c) we delete $p=75\%$ of the time-steps of the trajectories randomly. (d) Fragmented trajectories are reconstructed using a generalized version of deep autoencoders, called a Hadamard deep autoencoder. The root mean square error between the true and the reconstructed trajectories is computed to be 2.9.}
\label{fig1}
\end{figure}

In order to empirically illustrate the technical details that we presented so far, we utilize a video, available in \cite{Benedettelli2009}, of a perfectly coordinated swarm of four robots each mingling in two-dimensions, see Fig.~\ref{fig1}(a). We track the trajectories of the robots in this video using Kanade–Lucas–Tomasi (KLT) tracker \cite{lucas1981iterative,tomasi1991detection}, a widely used MOT method, with its default parameter values. We show the tracked trajectories of the robots by red, blue, yellow, and green in Fig.~\ref{fig1}(a). Now, we produce a neighbor graph of the trajectories; then, compute the shortest graph distances; finally, convert the shortest distance matrix to its Gramian matrix. Fig.~\ref{fig1}(b) exhibiting the SV percentages of the Gramian matrix ensures that the system has two major and many minor, degrees of freedoms since only two SVs are prominent while the rest of them are non-prominent. The total variance of these SVs is equal to the total dynamics described by this multi-agent system. These two SVs describe around $50\%$ of the systems' dynamics and 20 largest SVs describe almost all the dynamics the system. Thus, the nonlinear approximation for the rank of this trajectory dataset is around 20. This low-rank nature of collective motion is the essential property that guarantees the best performance of our HDA based trajectory reconstruction method. To mimic natural fragmentation of the trajectories, we randomly delete 75\% of the time-steps of the trajectories as shown in Fig.~\ref{fig1}(c).  We utilize HDA to reconstruct the fragmented trajectories where we observe a significantly better performance as the root mean square error (RMSE) between true and reconstructed trajectories is 2.9.

This paper is structured with four main sections: introduction, methods (Sec.~\ref{sec:method}), performance analysis  (Sec.~\ref{sec:perAna}), and discussion (Sec.~\ref{sec:diss}). In Sec.~\ref{sec:method}; first, we provide a mathematical background of the low-rank nature of collective motion; and second, we provide the formulation of both linear and nonlinear approximations to the rank; third, we present technical details of HDA in the context of trajectory completion; then, we present a LMC method \cite{lin2010augmented} for trajectory completion, as a platform for comparisons; finally, we present a generalized version of the Vicsek model \cite{Vicsek1995}, that we use to simulate data. Sec.~\ref{sec:perAna} contains an analysis of the performance of HDA against that of LMC using three diverse collective motion examples where two of them are simulated artificially using the classic Vicsek model \cite{Vicsek1995} and the other is a real-life robot swarm. We provide the conclusions along with a discussion in Sec.~\ref{sec:diss}. 

\begin{table}[htp]
\caption{Notations used in this paper and their descriptions}
\begin{tabular}{p{1.6cm}|p{11cm}}
Notation &  Description \\
\hline\\
$\zeta$ & Activation function.\\
$\eta$ & Epoch threshold. \\
$\tau$ & Error tolerance. \\ 
$i$ & Index for agents where $1\le i\le n.$\\
$t$ & Index for time-steps where $1\le t\le T.$\\
$\alpha$ & neighborhood parameter of graph. \\
$\epsilon^{(t)}_i$ & Noise parameter imposed to the $i$-th agent at the $t$-th time-step.\\
$\|\cdot\|_*$ & Nuclear norm (of a matrix).\\
$\theta^{(t)}_i$ & Orientation of the $i$-th agent at the $t$-th time-step.\\
$\mu$ & Penalty parameter of Lagrangian multiplier. \\
$p$ & Percentage of deletion of trajectories. \\
$r_d$ & Radius of interaction. \\
$N^{(t)}_i$ & $r_d$-radius of neighborhood of the $i$-th agent at the $t$-th time-step.\\
$\rho(\cdot)$ & Rank (of a matrix).\\
$\gamma^{(t)}$ & Rotation angle of the upward bell-shaped curve at the $t$-th time-step. \\
$\sigma_j$ & $j$-th singular value.\\
$\delta$ & Time-step size. \\

\hline
$\hat{\bs{\mathcal{A}}}^{(t)}$ &  ANN output of the configuration vector $\bs{\mathcal{A}}^{(t)}$.\\
$\bs{u}^{(t)}_i$ & Average direction of motion of the agents in the neighborhood of $N^{(t)}_i$.\\
$\bs{\mathcal{A}}^{(t)}$ &  Configuration vector of all the agent at the $t$-th time-step.\\
$\bs{\omega}^{(t)}$ & Indicator vector at the $t$-th time-step.\\
$\bs{\mathcal{A}}^{(t)}_k$ &  Input for the $k$-th layer of the ANN.\\
$\bs{c}^{(t)}$ & Point (2D) on the upward bell-shaped curve at the $t$-th time-step. \\
$\bs{a}^{(t)}_i$ &  Position vector (2D) of the $i$-th agent at the $t$-th time-step.\\
$\hat{\bs{a}}^{(t)}_i$ & Reconstructed position vector (2D) of the $i$-th agent at the $t$-th time-step.\\
$\bs{v}^{(t)}_i$ & Velocity of the $i$-th agent at the $t$-th time-step.\\

\hline\\
$\mathcal{H}$ & High-dimensional space of configurations.\\
$\mathcal{M}$ & Linear manifold.\\

\hline
$D$ & Decoder. \\
$E$ & Encoder. \\
$\Phi^{(t)}$ & Mapping on manifold at the $t$-th time-step.\\
$\mathcal{P}_{\Omega}$ & Projection over $\Omega$.\\
$S$ & Soft-threshold function. \\
$\mathbb{U}[a,b]$ & Uniform random distribution between $a$ and $b$.\\

\hline
$\mathcal{X}$ & Feature matrix representing the configuration of all of the agents.\\
$\mathcal{S}$=$\left[s^2_{tt'}\right]_{T\times T}$ & Gramian matrix. \\
$\Omega$ & Indicator matrix.\\
$L$ & Intermediate matrix in low-rank matrix completion optimization problem.\\
$\nu$ & Lagrangian multiplier. \\
$\mathcal{D}$=$\left[d^2_{tt'}\right]_{T\times T}$ & Squared Euclidean matrix. \\
$L$ & Reconstructed matrix by low-rank matrix completion.\\
$R^{(t)}_j$ & Rotation matrix of the $i$-th agent at the $t$-th time-step.\\
$U$ and $V$ & Unitary matrices.\\
$W_k$ & Weight matrix for the neurons between the layers $k-1$ and $k$.
\end{tabular}
\end {table}\label{tab:nom}

\subsection{Contributions}
Our fragmented trajectory reconstruction scheme presented in this paper makes the following contributions to the literature:
\begin{itemize}
\item Fragmented trajectories of collective motion are reconstructed using a deep autoencoder where the loss-function is defined only at observed segments of the trajectories by incorporating an indicator matrix into the loss-function with Hadamard product. 
\item Superior performance in fragmented collective motion trajectory reconstruction of our HDA is solely based on its nonlinear nature.
\item Collective motion trajectory data is nonlinear in nature.
\item The linear LMC method in \cite{lin2010augmented}, is capable of producing a faithful reconstruction similar to HDA until a high fragmentation level, but less accurate than HDA for further fragmentation.
\item HDA performs exceptionally well than LMC when fragmented trajectories of real-life collective motion are the data of interest.
\end{itemize}

\section{Methods}\label{sec:method}
This section provides technical details from data generation to the construction of the proposed trajectory completion method. First, we present the low-rank nature of collective motion in Sec.~\ref{sec:LRCM}. Second, in Sec.~\ref{sec:ranks}, we present two approaches, linear and nonlinear, of computing rank of trajectory data and justify of why HDA outperforms LMC in fragmented trajectory reconstruction. Third, we present the construction of HDA in the context of fragmented trajectory reconstruction in Sec.~\ref{sec:HDA}. Then, Sec.~\ref{sec:LMC} presents LMC that we use as the second method to compare the performance of HDA. Finally, Sec.~\ref{sec:vicsek} provides a generalized version of a classic particle simulator called Vicsek model.

\subsection{Low-rankness of Collective Motion}\label{sec:LRCM}
Low-rank nature of collective motion data grantees better performance of trajectory reconstruction schemes such as HDA and LMC; thus, here we discuss how low-rank is involved in collective motion data. For our study, we assume that the agents in collective motion move in a two-dimensional space. For a group of $n$ agents that traverses through $T$ total time-steps, the position vector of the $i$-th agent at the $t$-th time-step is denoted by $\bs{a}^{(t)}_i=\left[x^{(t)}_i,y^{(t)}_i\right]^{T_r} \in \mathbb{R}^2$, where $T_r$ denotes matrix transform. A vector representing the coordinates of all the agents at any time-step $t$ is called a \emph{configuration vector} of the agents at that time-step, which is defined as
\begin{equation}\label{eqn:configuration}
\mathcal{A}^{(t)}=\left[\bs{a}^{(t)}_1; \bs{a}^{(t)}_2; \dots; \bs{a}^{(t)}_n\right] \in \mathbb{R}^{2n\times 1}.
\end{equation}
This is a point on a high-dimensional Euclidean space where it represents the entire group at time $t$. The dataset of this study is made by concatenating all the $T$ configuration vectors vertically as
\begin{equation}\label{eqn:dataset}
\mathcal{X}=\left[\mathcal{A}^{(1)}, \mathcal{A}^{(2)}, \dots,  \mathcal{A}^{(T)}\right]\in \mathbb{R}^{2n \times T}.
\end{equation}
The dataset $\mathcal{X}$ can be represented as a high-dimensional data-cloud in Euclidean space, denoted as $\mathcal{H}$. This high-dimensional data-cloud can often be projected onto a \emph{low-dimensional linear subspace}, defined as $\mathcal{M}$, using the map
\begin{equation}\label{eqn:proj_mani}
\Psi:\mathcal{H}\rightarrow \mathcal{M} \ \ \mathrm{such \ that} \ \ \Psi\left(\mathcal{A}^{(t)}\right)=\mathcal{B}^{(t)},
\end{equation}
for some $\mathcal{B}^{(t)} \in \mathcal{M}$, \cite{gajamannage2016,Gajamannage2015}. This \emph{linear subspace} that describes the underlying structure of the data satisfies all the properties of being a \emph{linear manifold}, but we use the term \emph{linear subspace} to make it more general. Moreover, these configurations of coordinated agents evolve in time according to the mapping 
\begin{equation}\label{eqn:mapp_mani}
\Phi:\mathcal{M}\rightarrow \mathcal{M} \ \ \mathrm{such \ that} \ \ \Phi\left(\Psi\left(\mathcal{A}^{(t)}\right)\right)=\Psi\left(\mathcal{A}^{(t+1)}\right),
\end{equation}
\noindent \cite{gajamannage2016,Gajamannage2015}. Two maps in Eqns.~\eqref{eqn:proj_mani} and \eqref{eqn:mapp_mani} describe the relationship between configurations on a low-dimensional manifold. Existence of this low dimensional manifold guarantees that the trajectories of collective motion are low-rank as we will describe in a sequel. 

The number of non-zero SVs (see Def.~\ref{def:svd}) of the Gramian matrix of $\mathcal{X}$ relates to the dimensionality of this manifold.  A small number of non-zero SVs implies existence of a low-dimenional manifold on which the agents reside (see Def.~\ref{def:rank}). We have seen in Fig.~\ref{fig1}(b) that the data representing trajectories of the robot swam is low-rank since the Gramian matrix of the dataset possesses two largest SVs with around $50\%$ of variance and 20 largest SVs with almost $100\%$ of variance.

\begin{definition}\label{def:svd}
Consider that $\mathrm{Diag}(\sigma_1, \dots \sigma_{\min(n,m)})$ represents a diagonal matrix with the diagonal $(\sigma_1, \dots \sigma_{\min(n,m)})$. Let $L\in\mathbb{R}^{n\times m}$ be a real valued matrix, and $U_{m\times m}$ and $V_{n\times n}$ are two unitary matrices such that $U^TU=I$ and $V^TV=I$, respectively. Then, SV decomposition of $L$ is $L=U\Sigma V^T$, where $\Sigma_{m\times n}= \ $$\mathrm{Diag}(\sigma_1, \dots \sigma_{\min(n,m)})$. Here, for $j=1 ,\dots, \min(n,m)$, $\sigma_j$ represents $j$-th SV of $L$ such that $\sigma_j\ge \sigma_{j+1}$. 
\end{definition}

\begin{definition}\label{def:rank}
Let $\{\sigma_1, \dots \sigma_{\min(n,m)}\}$ be the SVs of the real valued matrix $L\in\mathbb{R}^{m\times n}$ as defined by Def.~\ref{def:svd}, then the rank of $L$, denoted by $\rho(L)$, is defined as the number of non-zero $\sigma_j$'s for $j=1,\dots,\min(m,n)$.
\end{definition}

Reconstruction of fragmented trajectories is primarily aided by the low-rank nature of the data generated from coordinated group motion. The rank of a dataset can be approximated using two approaches, called linear and nonlinear, where the reconstruction performance is heavily depend on whether the method in use is linear or nonlinear. Thus, we discuss these two aspects of the rank in the next section.

\subsection{Linear and nonlinear approximation of rank}\label{sec:ranks}
HDA is a nonlinear technique that we utilize for trajectory reconstruction whereas LMC is a linear technique for the same task. Since our focus is on trajectories extracted from coordinated group behavior (i.e., collective motion) of agents, such data can always be projected onto a low-dimensional manifold as described in Eqns.~\eqref{eqn:proj_mani} and \eqref{eqn:mapp_mani}. This ensures that the original high-dimensional trajectory data is low-rank where the prominent SVs along with their singular vectors of the Gramian matrix describe the underlying manifold. Since the rank of a dataset can be approximated using two approaches, linear and nonlinear, we provide technical details of those two approaches here. The linear approximation for the rank of a dataset is formulated as the number of non-zero SVs of the Gramian matrix of the squared Euclidean distance matrix of the dataset \cite{Gajamannage2021}. For that, first, we compute the squared Euclidean distance matrix between configurations as 
\begin{equation}
\mathcal{D}=\left[d_{t t'}^2\right]_{T\times T} \ ; \ \ d_{tt'} = \left\|\bs{\mathcal{A}}^{(t)} - \bs{\mathcal{A}}^{(t')} \right\|_2,
\end{equation} 
where $T$ is the number of points in the dataset (same as the total time-steps of a trajectory). Then, we transform the squared distance matrix $\mathcal{D}$ into its Gram matrix $\mathcal{S}=[s_{tt'}]_{T\times T}$, by employing \emph{double-centering} as
\begin{equation}\label{eqn:double_centering}
s_{tt'}=-\frac{1}{2}\big[d^2_{tt'}-\mu_t(d^2_{tt'}) -\mu_{t'}(d^2_{tt'})+\mu_{tt'}(d^2_{tt'})\big],
\end{equation}
where, $\mu_t(d^2_{tt'})$ and $\mu_{t'}(d^2_{tt'})$ are the means of the $t$-th row and $t'$-th column, respectively, of the squared distance matrix $\mathcal{D}$, and $\mu_{tt'}(d^2_{tt'})$ is the mean of the entire matrix $\mathcal{D}$  \cite{lee2007nonlinearb}. Finally, we compute the SV decomposition of $\mathcal{S}$ as Def.~\ref{def:svd} where the linear approximation for the rank is the number of non-zero SVs according to Def.~\ref{def:rank}. However, due to finite precision in computer arithmetic, the computed Gramian matrix $\mathcal{S}$ might be slightly deviated from being perfectly low-rank. This might results a few big SVs and significantly small rest of the SVs rather than having a few big SVs and zero rest of the SVs. 

Nonlinear approximation of rank of a matrix is formulated as an extension of the computation of the above linear approximation of rank where we replace the squared Euclidean distance matrix $\mathcal{D}$ with a squared geodesic distance matrix as similar to the dimensionality reduction technique presented in \cite{tenenbaum2000global}. For that, first, we create a graph structure, based upon configuration vectors $\left\{\bs{\mathcal{A}}^{(t)}\big\vert t = 1, \dots, T\right\}$ of the trajectories, where this graph estimates the intrinsic geometry of the nonlinear manifold as described by Eqns.~\eqref{eqn:mapp_mani} and \eqref{eqn:proj_mani}. For a given neighborhood parameter, say $\alpha$, we chose $\alpha$ nearest configurations to each configuration using Euclidean distance \cite{agarwal1999geometric}, and then transform the nearest neighbors into a graph structure by treating the configurations as graph nodes and connecting each pair of nearest neighbors by an edge having the weight equal to the Euclidean distance between them. The \emph{shortest path distances} in a graph are commonly considered as good approximations to the \emph{geodesic distances}. The shortest path distances between pairs of nodes in a graph can be computed using the Floyd's algorithm \cite{floyd1962algorithm}, as we presented in \cite{SGE}, which computes the shortest paths between all the pairs of points in one batch with less computational complexity. 

Similar to the linear approximation of rank, now we formulate the doubly centered matrix $\mathcal{S}$ from the squared geodesic distance matrix using Eqn.~(\ref{eqn:double_centering}). Since our computational process uses shortest graph distance to approximate the true geodesic distance, $\mathcal{S}$ might be slightly deviated from being perfectly low-rank. Thus, other than prominent SVs, the SV decomposition of the matrix $\mathcal{S}$ might produce significantly small SVs instead of SVs of zeros. The number of prominent SVs is defined as the nonlinear rank of the dataset.

The data of the fragmented trajectories is incomplete as the coordinates of missing trajectory segments do not present. Since such a dataset representing collective motion is low-rank, a dimensionality reduction technique such as DA \cite{Liou2014} can conveniently be used. In order to implement a conventional DA on trajectory datasets with missing entrees, we introduce the Hadamard product between the loss-function of the DA and an indicator matrix of ones and zeros as we discuss in the next section. 

\subsection{Hadamard Deep Autoencoder}\label{sec:HDA}
The fragmented trajectory reconstruction technique, HDA, that we present here is a generalized version of DAs. DAs are effectively used for learning low-dimensional representations of data since the encoder of DAs inputs high-dimensional data and then compresses that to low-dimensional representations using nonlinear functions, called activation functions. A DA learns a nonlinear map from the input data, denoted by $\mathcal{X}$, to the output, denoted by $\hat{\mathcal{X}}$, using two joined multi-layer ANNs where the first ANN performs encoding, denoted by $E$, and the other ANN connected to the end of the first one performs decoding, denoted by $D$. We assume that our trajectory data has $n$ agents evolving in a two-dimensional space for $T$ time-steps; thus, $\mathcal{X}$ and $\hat{\mathcal{X}} \in\mathbb{R}^{2n\times T}$. Since a row of $\mathcal{X}$ represents the same coordinate of an agent through all the time-steps, rows considered to be the features of the trajectory dataset. Since a column of $\mathcal{X}$ represents a configuration vector for a given time-step, columns considered to be data instances. The nonlinear map between input and output of HDA is given as
\begin{equation} \label{eq:map}
\begin{split}
E: \mathbb{R}^{2n\times T} \rightarrow \mathbb{R}^{2n'\times T} \ \text{and} \
D: \mathbb{R}^{2n'\times T} \rightarrow \mathbb{R}^{2n\times T}, \\
\text{such that} \ \hat{\mathcal{X}} = D\left(E\left(\mathcal{X}\right)\right),
\end{split}
\end{equation}
where $n\ge n'$ \cite{oftadeh2020eliminating}. Training a DA is equivalent to revealing the nonlinear sub-maps $E$ and $D$ that is implemented by minimizing the reconstruction error, also called the loss-function, such that
\begin{equation} \label{eq:opt1}
\mathop{\mathrm{arg min}}_{D,E} \|\mathcal{X}- D(E(\mathcal{X}))\|_F^2,
\end{equation}
where $F$ represents the Frobenius norm \cite{oftadeh2020eliminating}. 

We introduce a binomial indicator matrix, denoted by $\Omega$, of the same size as of $\mathcal{X}$, where it has ones at the locations of the observed entries of $\mathcal{X}$ and zeros at the locations of missing entries of $\mathcal{X}$. HDA incorporates this indicator matrix into the loss-function as,
\begin{equation} \label{eq:opt2}
\mathop{\mathrm{arg min}}_{D, E} \|\left(\mathcal{X}- D\left(E\left(\mathcal{X}\right)\right)\right) \odot \Omega \|^2_F,
\end{equation}
where $\odot$ represents the Hadamard or the element-wise product between the loss-function and $\Omega$. The key feature of this technique is that the training of the ANN is only based on the observed portion of the trajectories. This encoder generates a nonlinear low-dimensional representation of the high-dimensional input data  by filtering out coarse observable including noise. Then, the decoder $D$ converts back this low-dimensional representation to its output which is a new version of the high-dimensional input data. 
 
\begin{figure}[tp]
\centering
\includegraphics[width=1\textwidth]{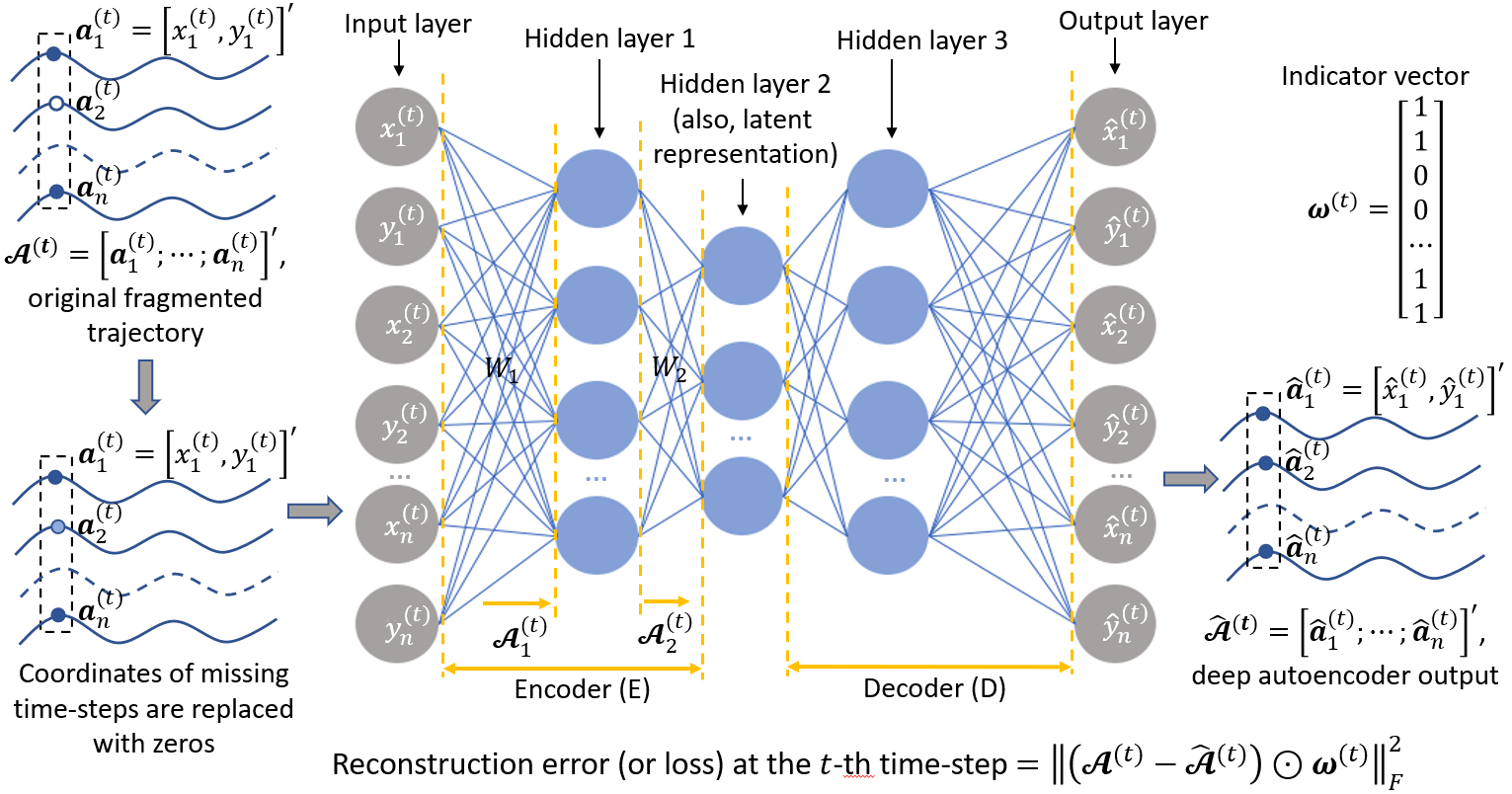}
\caption{An artificial neural network (ANN) model with three layers that is associated with the Hadamard deep autoencoder (HDA) for trajectory reconstruction. Unobserved entries of each configuration vector $\bs{\mathcal{A}}^{(t)}$, for $t=1,\dots,T$, are replaced with zeros and fed into the ANN of HDA. For the $t$-th time-step, observed, unobserved, and zero-filled entries on the trajectories are shown by disks in dark blue, white, and light blue, respectively. The indicator vector for $\bs{\mathcal{A}}^{(t)}$ of this model is $[1,1,0,0,\cdots,1,1]^{T_r}$ ($T_r$ denotes matrix transform) since the first agent is observed, the second agent is not observed, and the last agent is observed. The reconstruction error of the ANN is computed as the squared Frobenius norm of the element wise, e.i, Hadamard, product between the indicator vector and the difference of ANN's input to its output. Weight of each neuron is updated with the current best value at each iteration by minimizing the overall reconstruction error. In practical implementations of HDA, the number of layers and the nodes in each layer are to be customized to have better predictions.}
\label{fig2}
\end{figure}

HDA is trained with one configuration $\bs{\mathcal{A}}^{(t)}$ at a time; thus, to facilitate that concern, we generalize the HDA formulation given in Eqns.~\eqref{eq:map}, \eqref{eq:opt1}, and \eqref{eq:opt2}. Input to the ANN of the HDA propagates through each neuron and an activation function at each of the layer of the ANN as shown in Fig.~\ref{fig2}. For the $k$-th layer of the ANN that has a total of $K$ hidden layers (excluding the input and the output layers), we denote the input configuration as $\bs{\mathcal{A}}^{(t)}_k$, the output of that configuration as $\bs{\mathcal{A}}^{(t)}_{k+1}$, and the weights matrix associated with the incoming neurons as $W_k$. For given element-wise activation function $\zeta$ and bias vector for the $k$-th layer $\bs{b}_k$, the relationship between $\bs{\mathcal{A}}^{(t)}_k$ and $\bs{\mathcal{A}}^{(t)}_{k+1}$ is given by 
\begin{equation} \label{map1}
\bs{\mathcal{A}}^{(t)}_{k+1}=\zeta \left(W_k\bs{\mathcal{A}}^{(t)}_{k}+\bs{b}_k\right), k=1,\cdots,K+1,
\end{equation}
with $\hat{\bs{\mathcal{A}}^{(t)}}=\bs{\mathcal{A}}^{(t)}_{K+2}$ and $\bs{\mathcal{A}}^{(t)}=\bs{\mathcal{A}}^{(t)}_1$. Thus, the output of HDA, denoted as $\hat{\bs{\mathcal{A}}^{(t)}}$, is explicitly given as 
\begin{equation} \label{map2}
\hat{\bs{\mathcal{A}}^{(t)}}=\zeta \left(W_{K+1}\left(\zeta \left(W_{K}\left(\cdots \zeta\left( W_1\bs{\mathcal{A}}^{(t)}+\bs{b}_1\right) \cdots\right)+\bs{b}_{K}\right)\right)+\bs{b}_{K+1}\right),
\end{equation}
and the reconstruction error, denoted as $L$, is
\begin{equation} \label{loss1}
L=\left\|\left(\bs{\mathcal{A}}^{(t)}-\hat{\bs{\mathcal{A}}^{(t)}}\right)\odot \omega^{(t)}\right\|^2_F,
\end{equation}
where $\omega^{(t)}$ is the indicator vector for the $t$-th time-step with ones and zeros corresponding to observed entries and unobserved entries, respectively, of the configuration vector $\bs{\mathcal{A}}^{(t)}$.

For given trajectories of collective motion evolving in $T$-time steps; first, we extract the configuration vectors $\bs{\mathcal{A}}^{(t)}$ for $t=1,\dots,K$. Then, we replace unobserved entries of each $\bs{\mathcal{A}}^{(t)}$ with zeros and feed that into the ANN of HDA to generate its output $\hat{\bs{\mathcal{A}}^{(t)}}$. The loss function, defined in Eqn.~\eqref{loss1}, is minimized using backpropagation \cite{goodfellow2016deep} to determine optimum weights $W_1,\dots,W_{K+1}$. After training the HDA with all the configuration vectors ($\bs{\mathcal{A}}^{(t)}$'s), we shuffle the rows of $\mathcal{X}$ while keeping each consecutive pair of rows, e.i., $(2i-1,2i)$ for $i=1,2,\dots,2n$, as a single unit. For the $i$-th agent, since the rows $2i-1$ and $2i$ represent $x$ and $y$ coordinates, respectively, such shuffling is the same as switching the order of the agents in the matrix $\mathcal{X}$. We train the same ANN with the new configuration vectors repeatedly until either the number of epochs reaches its maximum, defined as $\eta$, or the reconstruction error meets the error tolerance, defined as $\tau$. After the training is completed, we feed each configuration vector $\bs{\mathcal{A}}^{(t)}$ into the ANN and compute the corresponding output $\hat{\bs{\mathcal{A}}^{(t)}}$. The reconstructed trajectories in a matrix form is given as $\hat{\mathcal{X}}=\left[\hat{\bs{\mathcal{A}}^{(1)}}, \cdots, \hat{\bs{\mathcal{A}}^{(t)}}, \cdots, \hat{\bs{\mathcal{A}}^{(T)}}\right]$. Algorithm~\ref{algo:HDA}(a) and Algorithm~\ref{algo:HDA}(b) show the procedures of training an ANN for HDA with fragmented trajectories and reconstruction of the fragmented trajectories using this trained ANN, respectively. Our code is developed in Keras \cite{chollet2015keras} with Tensorflow backend \cite{tensorflow2015-whitepaper}.

\begin{algorithm}[htp]
\caption{HDA for the reconstruction of fragmented trajectories.}
\label{algo:HDA}
\SetAlgoLined
\textbf{(a) Training}\\
\KwIn{Partially observed trajectory matrix, $\mathcal{X}=\left[\bs{\mathcal{A}}^{(1)}, \cdots, \bs{\mathcal{A}}^{(t)}, \cdots, \bs{\mathcal{A}}^{(T)}\right]$; indicator matrix, $\Omega=\left[\bs{\omega}^{(1)},\cdots,\bs{\omega}^{(t)},\cdots,\bs{\omega}^{(T)}\right]$; activation function, $\zeta$; number of maximum epochs, $\eta$; and error tolerance, $\tau$.}
\KwOut{Optimized weight matrices, $W_1,\dots,W_{K+1}$.}
 Initialization: Index $j=1$; generate a zero matrix, denoted as $\hat{\mathcal{X}}_1$, of the size of $\mathcal{X}$; weight matrices, $\{W_k| \ \text{for all} \ k\}$, are set to zeroes. \\
 \While{$\left\| \left(\mathcal{X} -  \hat{\mathcal{X}}_j\right) \odot \Omega\right\|_F \ge \tau$ AND $j\le \eta$}{
  Shuffle $\bs{\mathcal{A}}^{(t)}$'s in $\mathcal{X}$ randomly. Then, shuffle both $\hat{\bs{\mathcal{A}}^{(t)}}$'s in $\hat{\mathcal{X}}_j$ and $\bs{\omega}^{(t)}$'s in $\Omega$ with the same order as $\bs{\mathcal{A}}^{(t)}$'s.\\
  \For{all $\bs{\mathcal{A}}^{(t)} \in \mathcal{X}$} {  	
  	Compute the reconstruction error (e.i, loss), $L=\left\|\left(\bs{\mathcal{A}}^{(t)}-\hat{\bs{\mathcal{A}}^{(t)}}\right)\odot \bs{\omega}^{(t)}\right\|^2_F$, of the ANN.\\
	Minimize $L$ with backpropagation to find the optimum weights, $W_1,\dots,W_{K+1}$. \\
	Update the ANN with the optimized weights. \\   	
	Compute the output, denoted as $\hat{\bs{\mathcal{A}}^{(t)}}$, of the ANN for $\bs{\mathcal{A}}^{(t)}$ using Eqn.~\eqref{map2}.\\
  }
	Produce $\hat{\mathcal{X}}_j=\left[\hat{\bs{\mathcal{A}}^{(1)}}, \cdots, \hat{\bs{\mathcal{A}}^{(t)}}, \cdots, \hat{\bs{\mathcal{A}}^{(T)}}\right]$. \\
 	$j = j + 1$
 }
\hrulefill\\
\textbf{(b) Reconstruction}\\
\SetAlgoLined
\KwIn{Partially observed trajectory matrix, $\mathcal{X}=\left[\bs{\mathcal{A}}^{(1)}, \cdots, \bs{\mathcal{A}}^{(t)}, \cdots, \bs{\mathcal{A}}^{(T)}\right]$; optimized weights, $W_1,\dots,W_{K+1}$, using Algorithm~\ref{algo:HDA}(a); and activation function, $\zeta$.}
\KwOut{Reconstructed trajectories, $\hat{\mathcal{X}}=\left[\hat{\bs{\mathcal{A}}^{(1)}}, \cdots, \hat{\bs{\mathcal{A}}^{(t)}}, \cdots, \hat{\bs{\mathcal{A}}^{(T)}}\right]$.} 
  \For{all $\bs{\mathcal{A}}^{(t)} \in \mathcal{X}$} {  
	Compute the output, $\hat{\bs{\mathcal{A}}^{(t)}}$, of the ANN for $\bs{\mathcal{A}}^{(t)}$ using Eqn.~\eqref{map2}.\\	
  }
Produce $\hat{\mathcal{X}}=\left[\hat{\bs{\mathcal{A}}^{(1)}}, \cdots, \hat{\bs{\mathcal{A}}^{(t)}}, \cdots, \hat{\bs{\mathcal{A}}^{(T)}}\right]$. \\
\end{algorithm}

\subsection{Low-rank Matrix Completion}\label{sec:LMC}
We compare the performance of HDA against a generalized version of the benchmark low-rank matrix completion (LMC) technique given in \cite{lin2010augmented}. We presented in our previous work \cite{Jayasumana2018network, Mahindre2019}, that LMC exhibit superior performance in topology recovery of both partially observed directed graphs and partially observed uncorrected graphs. In a similar vein, we have also shown that LMC methods have significant performance in reconstruction of fragmented trajectories because the coordinate matrix representing agents’ trajectories of collective motion is often low-rank due to mutual interactions and dependencies between the agents \cite{Gajamannage2019Reconstruction}. In this context, the input to this LMC method is a partially observed coordinates matrix where the output consists the fully observed portion of the original matrix as well as the approximated values for the unobserved portion. This LMC method guarantees that the observed segments of the trajectories in the input coordinates matrix $\mathcal{X}$ remain the same as those in the reconstructed dataset. 

Let, matrix $L\in\mathbb{R}^{2n\times T}$ represents the reconstructed trajectories at an intermediate iteration of the LMC scheme in \cite{lin2010augmented}.  The constraint $\mathcal{P}_{\Omega}(\mathcal{X}) = \mathcal{P}_{\Omega}(L)$ imposes the observed entries in the trajectory matrix $\mathcal{X}$ remain the same in the intermediate reconstructed matrix $L$. Therefore, the optimization scheme associated with the construction of the recovered matrix $\hat{\mathcal{X}}$ of $\mathcal{X}$ is
\begin{equation}\label{eqn:MC}
\hat{\mathcal{X}} = \arg \min_{L} \rho(L), \ \ 
\text{s.t.}\; \mathcal{P}_{\Omega}(\mathcal{X}) = \mathcal{P}_{\Omega}(L),
\end{equation}
where $\rho(\cdot)$ is the rank operator as in Def.~\ref{def:rank}. That is to say, we seek to find a matrix $L$ such that the rank of $L$ [denoted by $\rho(L)$] is minimized by enforcing the constraint that the observed entries in $\mathcal{X}$ matches the corresponding entries of the reconstructed matrix $\hat{\mathcal{X}}$. Thus, the reconstructed matrix $L$ will retain the observed portion of the trajectories same, but $L$ will take any values for the unobserved portion to minimize the rank of $L$.

The optimization problem in Eqn.~\eqref{eqn:MC} is non-convex
\footnote{A function $f$, defined on the domain $D$, is said be convex if for all $x, y \in D$  and all $\theta \in [0,1]$, it holds both $\theta x + (1-\theta)y \in D$ and $f(\theta x + (1-\theta)y)\le\theta f(x) + (1-\theta)f(y)$.}
 and NP-hard
\footnote{A problem is NP-hard if an algorithm for solving it can be translated into an algorithm for solving any non-deterministic polynomial time problem.}; 
thus, we obtain a convex surrogate to that by replacing the rank operator by nuclear norm as
\begin{equation}\label{eqn:MCconvex}
\hat{\mathcal{X}} = \arg \min_{L} \| L \|_*, \ \ 
\text{s.t.}\; \mathcal{P}_{\Omega}(\mathcal{X}) = \mathcal{P}_{\Omega}(L),
\end{equation}
where $\| L \|_*$ is the nuclear norm of $L$ as defined in Def.~\ref{def:nuclear}. The constrain in Eqn.~\eqref{eqn:MCconvex} is implemented using a soft-threshold function $S\in\mathbb{R}^{2n\times T}$ where
\begin{equation}\label{eqn:thresh}
\begin{aligned}
S[i,j] = \left\{ \begin{array}{cc} 
                0 & ; if \Omega[i,j] = 0, \\
                \left\vert \mathcal{X}[i,j]-L[i,j]\right\vert & ; if \Omega[i,j] = 1.  \\            
                \end{array} \right.
\end{aligned}
\end{equation}
Thus, the optimization scheme in Eqn.~\eqref{eqn:MCconvex} is modified as
\begin{equation}\label{eqn:MCconvex1}
\hat{\mathcal{X}} = \arg \min_{L} \| L \|_*, \ \ 
\text{s.t.}\; S = 0.
\end{equation}
As an intermediate step of solving the optimization problem in Eqn.~\eqref{eqn:MCconvex1}, we write the augmented Lagrangian function of this optimization problem with Lagrangian multiplier $\nu\in \mathbb{R}^{2n\times T}$ and penalty parameter $\mu\in \mathbb{R}$ such that
\begin{equation}\label{eqn:MCconvex2}
\mathcal{L}(L,\nu) = \| L \|_* + \langle \nu,S\rangle + \frac{\mu}{2} \|\nu\|^1_F,
\end{equation}
where $\langle \cdot,\cdot \rangle$ represents Frobenius inner product as defined in Def.~\ref{def:innerP}. The value for the penalty parameter $\mu$ is estimated to be $1.25/\|\mathcal{P}_{\Omega}(\mathcal{X})\|_F$ as per guidance in \cite{lin2010augmented}. Then, we use Alternating Direction Method of Multipliers (ADMM) \cite{boyd2010distributed}, to optimize $\mathcal{L}(L,\nu)$ in Eqn.~\eqref{eqn:MCconvex2} by one variable at a time. Algorithm \ref{algo:MC} shows the procedure for LMC that reconstructs the missing segments of the trajectories through the augmented Lagrangian function and ADMM optimization.

\begin{definition}\label{def:nuclear}
Let, $\{\sigma_1, \dots \sigma_{\min(n,m)}\}$ be the SVs of the real valued matrix $L\in\mathbb{R}^{m\times n}$ as defined by Def.~\ref{def:svd}. Then, the nuclear norm of $L$, denoted by $\|L\|_*$, is defined as $\sum_{j=1}^{\min(n,m)} \sigma_j$.
\end{definition}

\begin{definition}\label{def:innerP}
Frobenius inner product of two matrices $A = [a_{ij}]_{m\times n}\in \mathbb{R}^{m\times n}$ and $B = [b_{ij}]_{m\times n}\in \mathbb{R}^{m\times n}$, denoted by $\langle A, B \rangle$, is defined as
\begin{equation}
\langle A, B \rangle = \sum_{1\le i \le m, 1\le j \le n} a_{ij}b_{ij}.
\end{equation}
\end{definition}

\begin{algorithm}[h]
\caption{LMC for the reconstruction of fragmented trajectories}
\label{algo:MC}
\SetAlgoLined
\KwIn{Partially observed trajectory matrix, $\mathcal{X}$; indicator matrix, $\Omega$; number of maximum iterations, $\eta$; and error tolerance, $\tau$.}
\KwOut{Reconstructed trajectories,  $\hat{\mathcal{X}}$ }
 	Initialization: $L=0$, $\nu=0$, $\mu=1.25/\|\mathcal{P}_{\Omega}(\mathcal{X})\|_F$ (according to \cite{lin2010augmented}), and $S=0$ (by Eqn.~\eqref{eqn:thresh} with $L=0$).\\
 \While{$\|S\|_F \ge \tau$ AND $j\le \eta$}{	
	Solve $\hat{\mathcal{X}}= \arg \min_{L} \mathcal{L}(L,\nu)$ in Eqn.~\eqref{eqn:MCconvex2} and update $L$ with $\hat{\mathcal{X}}$. \\
	Solve $\hat{\nu}= \arg \min_{\nu} \mathcal{L}(L,\nu)$ in Eqn.~\eqref{eqn:MCconvex2} and update $\nu$  with $\hat{\nu}$. \\	
	Compute $S$ with the current $L$ using Eqn.~\eqref{eqn:thresh}. \\
	j = j + 1 \\  
}
\end{algorithm}

\subsection{Vicsek model}\label{sec:vicsek}
This study is carried out using three datasets where two of them are collective motion simulations and the other one is a real-life instance. Collective motion is defined as the spontaneous emergence of ordered movement in a system consisting of many self-propelled agents, and one of the most famous models that describes such behavior is the Vicsek model given in \cite{Vicsek1995}, which performs simulations on a square shaped region with periodic boundary conditions. We generate the two synthetic collective motion datasets using a generalized version of the classic Vicsek model that we modified with a rotational matrix \cite{Gajamannage2019Reconstruction}. This agent-wise temporal rotation matrix imposed to the $i$-th agent at the $t$-th time-step is denoted as $R^{(t)}_i$ which allows us to simulate interesting collective motion scenarios while ensuring the intra-group interactions between the agents. 

The Vicsek model updates the orientation of the $i$-th particle at the $t$-th time-step, denoted by $\theta_i^{(t)}\in[-\pi, \pi]$, based on the average direction of motion of all the particles in a neighborhood, denoted by $N^{(t)}_{i}$, within a radius $r_d$ of the $i$-th particle.  Thus, the orientation of the $i$-th agent at the $t$-th time-step is defined as $\theta_i^{(t+1)}$ that we compute using
\begin{equation}\label{eqn:modVicsek1}
\theta_i^{(t+1)}=\arg(\bs{u}^{(t)}_i)+\epsilon_i^{(t)};
\end{equation}
where, $\epsilon_i^{(t)}$ is a noise parameter imposed to the orientation of the i-th agent at the t-th time-step, and 
\begin{equation}\label{eqn:modVicsek}
\bs{u}^{(t)}_i=\frac{1}{|N^{(t)}_{i}|} \sum_{j\in N^{(t)}_{i}} R^{(t)}_j \begin{bmatrix}\cos(\theta_j^{(t)}) \\ \sin(\theta_j^{(t)})\end{bmatrix}.
\end{equation}
The position vector of the $i$-th agent at the $t$-th time-step is given as
\begin{equation}\label{eqn:modVicsek2}
\bs{a}_i^{(t+1)}=\bs{a}_i^{(t)}+v_i^{(t)}R_i^{(t)} \begin{bmatrix}\cos(\theta_i^{(t)}) \\ \sin(\theta_i^{(t)})\end{bmatrix} \delta; \\
\end{equation}
where, $v_i^{(t)}$ denotes the speed of the $i$-th agent at the $t$-th time-step and $\delta$ is the step size. Eqns.~\eqref{eqn:modVicsek1}, \eqref{eqn:modVicsek}, and \eqref{eqn:modVicsek2} with $R_i^{(t)}=[1,0;0,1]$ for all $i$'s and $t$'s represent the classic Vicsek model. In Sec.~\ref{sec:perAna}, we will simulate a spiral collective motion scenario and a collective motion scenario at an obstacle using this generalized Vicsek model.

\section{Performance Analysis} \label{sec:perAna}
Here, we use two collective motion datasets produced by the generalized Vicsek model and another dataset produced by tracking trajectories of a real-life collective motion videos. The two synthetic datasets represent one spiral group motion and a group motion at an obstacle, whereas the two real-life datasets represent one mingling robot swarm and one free style school of fish.

\subsection{Spiral collective motion}\label{sec:spiral}
Since spiral collective motion is often observed in the nature ranging from biology, such as bacteria colonies \cite{Koizumi2020}, to astronomy, such as spiral galaxies \cite{Parnovsky2010}, we simulate and analyze trajectories representing a spiral collective motion scenario. First, we formulate an anticlockwise Archimedean spiral that rotates an angle of $3\pi$ and then compute the agent-wise temporal rotational matrix based on that spiral. We assume that the two-dimensional coordinates $\left[\bs{c}^{(1)}; \dots; \bs{c}^{(t)}; \dots; \bs{c}^{(T)}\right]$ where $\bs{c}^{(t)} \in\mathbb{R}^2$ represent this spiral such that
\begin{equation}
\begin{split}
\bs{c}^{(t)}=r^{(t)} \begin{pmatrix} \cos\left(\kappa^{(t)}\right) \\ \sin\left(\kappa^{(t)}\right) \end{pmatrix} \ ; \ \text{where} \\
r^{(t)} = 1+\frac{3}{T-1}(t-1) \ \text{and} \ \kappa^{(t)}=\frac{3\pi}{T}(t-1),
\end{split}
\end{equation}
for $t=1, \dots, T$. Here, $r^{(t)}$ is the variable radius of the spiral where it changes from 1 to 4 and $\kappa^{(t)}$ is the angle of rotation with respect to the origin that varies from 0 to $3\pi$. The rotation angle with respective to the $(t-1)$-th coordinates, $\bs{c}^{(t-1)}=\left[c^{(t-1)}_1, c^{(t-1)}_2\right]^{T_r}$, of this spiral, denoted by $\gamma^{(t)}$, is
\begin{equation}
\gamma^{(t)}=\tan^{-1}\left(\frac{c_2^{(t)}-c_{2}^{(t-1)}}{c_1^{(t)}-c_{1}^{(t-1)}}\right) \ ; \ t=2,\dots,T.
\end{equation} 
Thus, the two-dimensional rotational matrix for the $i$-th agent at the $t$-th time step is given by
\begin{equation}\label{eqn:rotmat1}
R^{(t)}_{i}=  
\begin{pmatrix} \cos\left(-\gamma^{(t)}\right) & -\sin\left(-\gamma^{(t)}\right) \\ \sin\left(-\gamma^{(t)}\right) & \cos\left(-\gamma^{(t)}\right) \end{pmatrix}
\end{equation}
Eqns.~\eqref{eqn:modVicsek1}, \eqref{eqn:modVicsek}, and \eqref{eqn:modVicsek2} with the rotational matrix in Eqn.~\eqref{eqn:rotmat1} provide the complete formulation of the system generating the spiral collective motion dataset.

\begin{figure}[htp]
\centering
\includegraphics[width=1\textwidth]{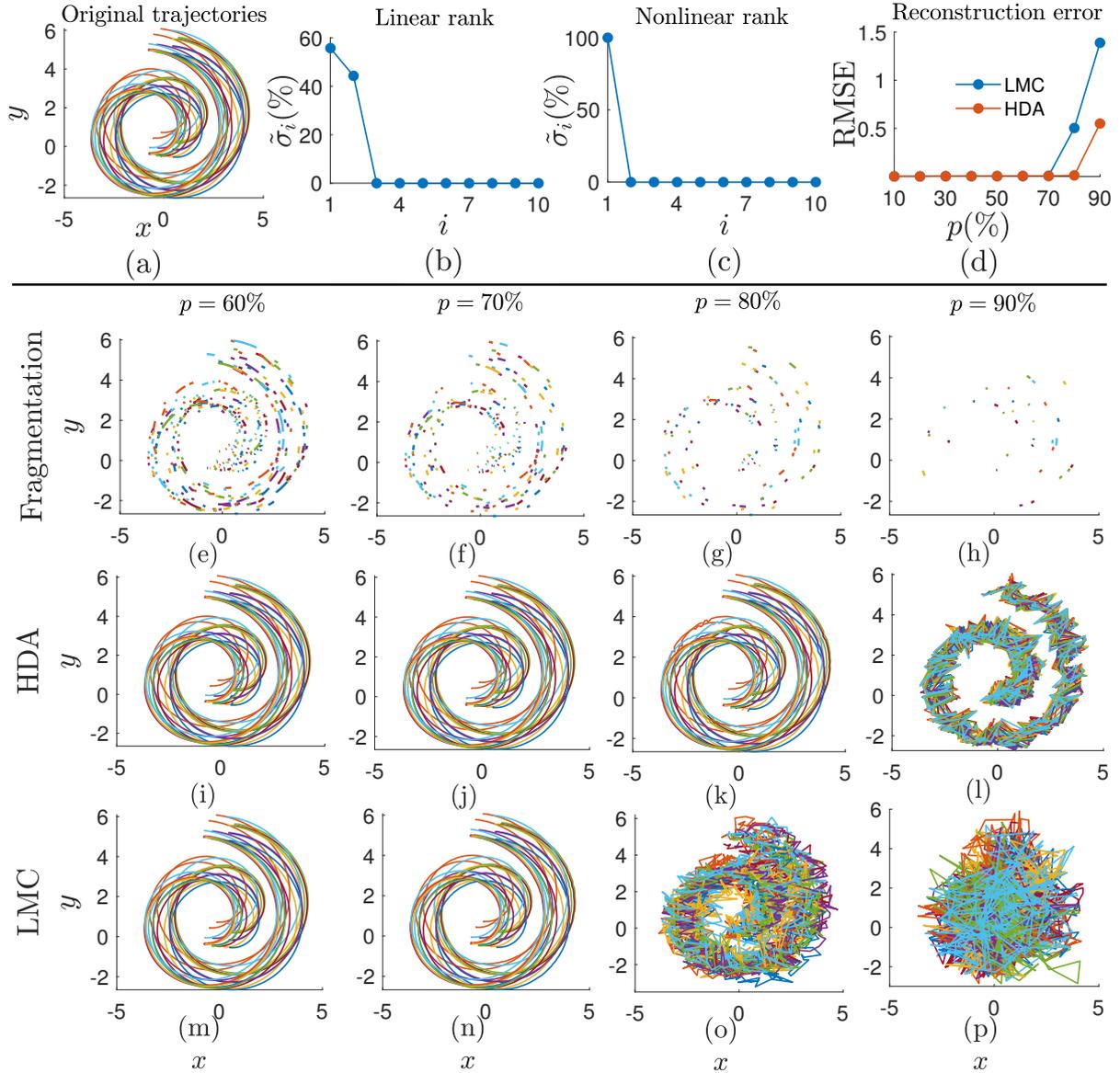}
\caption{Reconstruction of fragmented trajectories of spiral collective motion. We generate, (a) a spiral collective motion dataset using a generalized Vicsek model. A linear and a nonlinear approximations for the rank of this dataset is computed where the percentages of the singular values (SVs) associated with those are shown in (b) and (c), respectively. Observation of two prominent SVs in (b) implies that the linear approximation of the rank is two, whereas the observation of one prominent SV in (c) implies the nonlinear approximation of the rank is one (the rank is approximated as the number of non-zero SVs). Nine datasets are generated by deleting variable percentages, say $p\%$'s, of the trajectories such that $p=10\%, 20\%, \dots, 90\%$ where the last four cases are shown in (e)--(h). Fragmented trajectories are reconstructed using HDA with the ReLu activation function, 1000 epochs, and the error tolerance of $10^{-6}$. Then, we reconstruct the same fragmented trajectories using LMC with 1000 iterations and the error tolerance of $10^{-6}$. (d) RMSE between the reconstructed and the original trajectories shows that both methods perform equally well when $p\le70$; however, RMSE of LMC increases rapidly than that of HAD when $p>70\%$. For the fragmentations $p=60\%$--$90\%$, (i)--(l) show the corresponding reconstructions generated by HDA and (m)--(p) show the corresponding reconstructions generated by LMC.}
\label{fig3}
\end{figure}

We generate a swarm of 20 agents ($n$), see Fig.~\ref{fig3}(a), traversing through 200 time-steps ($T$) in a rectangular domain with periodic boundary conditions. Moreover, we set the radius of interaction ($r_d$) to 1, the speed of each agent at each time-step ($v_i^{(t)}$ for all $i$ and $t$) to .05, the time-step size ($\delta$) to 1, and noise of each particle at each time-step ($\epsilon_i^{(t)}$ for all $i$ and $t$) to .01. As we stated early, HDA is a nonlinear technique to reconstruct fragmented trajectories, whereas LMC is a linear technique that we utilize to perform the same reconstruction task. Thus, we approximate the rank of this spiral dataset both linearly and nonlinearly as explained in Sec.~\ref{sec:ranks}. Consider that the SV spectrum corresponds to the rank approximation is $\left\{\sigma_i | 1 \le i \le \min(T,2n)\right\}$, then we define the percentage of the $i$-th SV, denoted by $\tilde{\sigma}_i$, as
\begin{equation}\label{eqn:per_SV}
\tilde{\sigma}_i = \frac{\sigma_i}{\sum_{i=1}^{\min(T,2n)} \sigma_i} 100\%.
\end{equation}
Fig.~\ref{fig3}(b) presents the percentages of SVs for the linear approximation of the rank while Fig.~\ref{fig3}(c) presents the percentages of SVs for the nonlinear approximation of the rank. We observed that while the linear approximation of the rank of the dataset is two and nonlinear approximation of the rank is one. Nonlinear approximation is a faithful approximation to the dataset's natural rank than its linear approximation if the two approximations are different. Given that the two rank approximations are different in this dataset, nonlinear machine learning methods such as HDA may perform well in reconstruction of trajectories whereas nonlinear machine learning methods such as LMC may not do the same. To emphatically justify this conjecture, we use both LMC and HDA for the reconstruction of the fragmented trajectories. 

We fragment the trajectories by deleting some segments of them at variable percentages, say $p\%$, of the time-steps such that $p=10\%,\dots,90\%$, where we show the last four fragmentations in Figs.~\ref{fig3}(e--h). Here, if we delete the $i$-th agent's trajectory at the $t$-th time-step, we delete both of the coordinates of that agent to mimic real-life occlusion caused by single camera recording. We produce the binary indicator matrix $\Omega$ that represents deleted time-steps of the data by zeros. We set the maximum epochs ($\eta$) to 1000, the error tolerance ($\tau$) to $10^{-6}$, and the activation function to ReLU, and train the HDA presented in Alg.~\ref{algo:HDA}(a). The trained weights $\left\{W_k | \ k=1, \dots, K\right\}$ are fed into Alg.~\ref{algo:HDA}(b) along with the indicator matrix $\Omega$ to obtain the reconstructed trajectories $\hat{\mathcal{X}}=\left[\hat{\bs{\mathcal{A}}^{(1)}}, \dots, \hat{\bs{\mathcal{A}}^{(t)}}, \dots, \hat{\bs{\mathcal{A}}^{(T)}}\right]$. We reconstruct the fragmented trajectories again by running the LMC scheme presented in Alg.~\ref{algo:MC} with 1000 iterations and $10^{-6}$ of error tolerance. 

We compute RMSE between the original trajectories, $\left\{\bs{\mathcal{A}}^{(t)} \big\vert \ t=1, \dots, T\right\}$, and the reconstructed trajectories, $\left\{\hat{\bs{\mathcal{A}}^{(t)}} \Big\vert \ t=1, \dots, T\right\}$, as 
\begin{equation} \label{eq:rmse}
RMSE =\sqrt{\frac{1}{T} \sum_{1\le t\le T}\left(\hat{\bs{\mathcal{A}}^{(t)}}-\bs{\mathcal{A}}^{(t)}\right)^2},
\end{equation} 
for each fragmentation level, that we show in Fig.~\ref{fig3}(d). We observe therein that both of the methods perform equally well when $p\le70\%$; however, HAD outperforms when $p>70\%$. As a visual aid, for the fragmentations $p=60\%$--$90\%$, we show the reconstructed trajectories in Figs.~\ref{fig3}(i--l) for HDA and in Figs.~\ref{fig3}(m--p) for LMC. These figures justify that HDA performs well until $80\%$ of fragmentation whereas LMC performs well only until $70\%$ of fragmentation. Since the linear rank of this dataset is different to the nonlinear rank of it, the linear technique LMC is only capable of learning a small portion of the underlying pattern of the data whereas HDA is capable of learning a bigger portion of the pattern. This discrepancy between two schemes makes diverse impacts on the performance of trajectory reconstruction .

\subsection{Collective motion at an obstacle}\label{sec:obstacle}
Analyzing collective motion instances at obstacles to learn dynamics of the animal groups is an interesting topic in science \cite{Xue2008}. Thus, we simulate a dataset that represents a collective motion scenario at an obstacle where a set of agents evolves as a single group at the beginning; then, they split into two groups to avoid an obstacle on their way; finally, they get together to form a single group after the obstacle is passed. This specific collective motion is simulated in few steps: first, we formulate an upward bell-shaped curve; second, we calculate the rotation angles of a particle traversing on this curve; third, we calculate the rotation angles of a particle traversing on a similar, but downward, bell-shaped curve; fourth, we randomly split the group into two portions and use these rotational angles such that one portion's centroid traverses through the upward bell-shaped curve while the other portion's centroid traverses on the downward bell-shaped curve.

Let, two-dimensional coordinates $\left[\bs{c}^{(1)}; \dots; \bs{c}^{(T)}\right]$, where $\bs{c}^{(t)}=\left(c_1^{(t)},c_2^{(t)}\right)\in\mathbb{R}^2$, represent the aforesaid upward bell-shaped curve. 
By discretization of the horizontal axis into $T$ segments ranging from $-6$ to 6, the first coordinates of the upward bell-shaped curve becomes 
\begin{equation}
c^{(t)}_1=6(2t-T-1)/(T-1) \ ; \ t=1, \dots, T.
\end{equation}
The second coordinates of this upward bell-shaped curve is formulated using difference of two sigmoid functions of the form $1/\left(1+e^{-(at-b)}\right)$, where $a$ and $b$ are constants, such that
\begin{equation}
c^{(t)}_2=5\left(\frac{1}{1+e^{-(\frac{T}{12}t-4)}}-\frac{1}{1+e^{-(\frac{T}{12}t-8)}}\right); t=1,\dots, T.
\end{equation}
Temporal rotation angles of this upward bell-shaped curve at the $t$-th time-step, denoted by $\gamma^{(t)}$, is
\begin{equation}
\gamma^{(t)}=\tan^{-1}\left((c_2^{(t)}-c_{2}^{(t-1)})\big/(c_1^{(t)}-c_{1}^{(t-1)})\right) \ ; \ t=2,\dots,T.
\end{equation} 
The temporal rotation angles of the downward bell-shaped curve is obtained by changing the sign of the rotation angles of the upward bell-shaped curve, that is as $-\gamma^{(t)}$ for $t=2,\dots,T$. Now, we split the entire particle swarm of $n$ agents into two groups such that
\begin{equation}\label{eqn:rotmat2}
R^{(t)}_{i}=  
\begin{cases}
\begin{pmatrix} \cos\gamma^{(t)} & -\sin\gamma^{(t)} \\ \sin\gamma^{(t)} & \cos\gamma^{(t)} \end{pmatrix}, & \text{if } \ \ 1 \le i \le \big\lfloor \frac{n}{2} \big\rfloor, \\
\begin{pmatrix} \cos(-\gamma^{(t)}) & -\sin(-\gamma^{(t)}) \\ \sin(-\gamma^{(t)}) & \cos(-\gamma^{(t)}) \end{pmatrix}, & \text{if } \ \  \big\lfloor \frac{n}{2} \big\rfloor+1 \le i \le n.
\end{cases}
\end{equation}
This setup makes each agent in the range $[1, \lfloor n/2 \rfloor]$ follows an upward bell-shaped pattern while each agent in the range $[\lfloor n/2 \rfloor+1, n]$ follows a downward bell-shaped pattern. We incorporate the rotational matrix in Eqn.~\eqref{eqn:rotmat2} with Eqns.~\eqref{eqn:modVicsek1}, \eqref{eqn:modVicsek}, and \eqref{eqn:modVicsek2} to produce the dataset of a collective motion scenario at an obstacle. 

\begin{figure}[htp]
\centering
\includegraphics[width=1\textwidth]{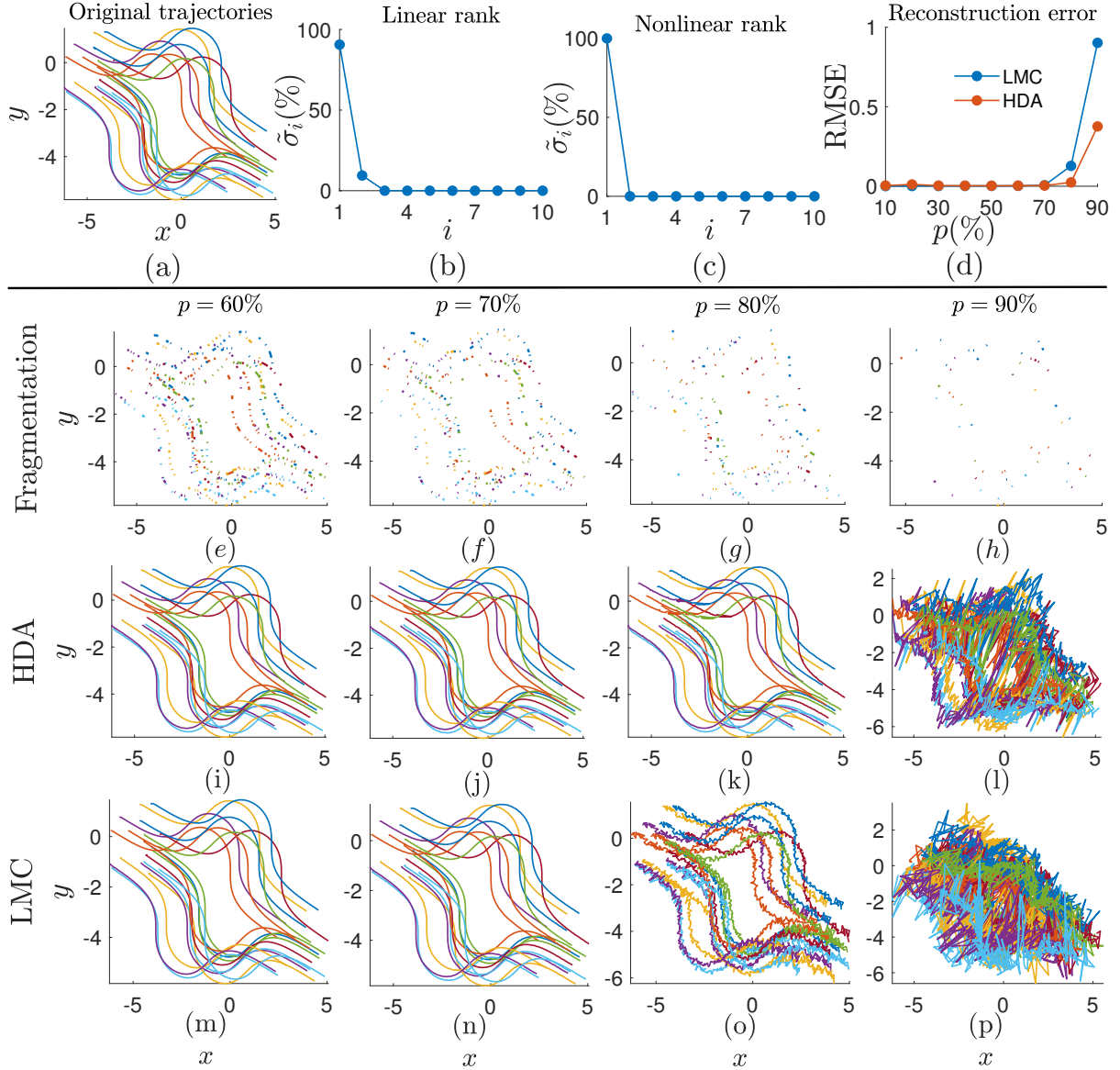}
\caption{Reconstruction of fragmented trajectories of collective motion at an obstacle. We generate, (a) a collective motion dataset to represent dynamics of agents at an obstacle using a generalized Vicsek model. A linear and a nonlinear approximations  for the rank of this dataset is computed where the percentages of the singular values (SVs) associated with those are shown in (b) and (c), respectively. Observation of two prominent SVs in (b) implies that the linear approximation of the rank is two, whereas the observation of one prominent SV in (c) implies the nonlinear approximation of the rank is one (the rank is approximated as the number of non-zero SVs). We generate nine datasets by deleting variable percentages, say $p\%$'s, of the trajectories such that $p=10\%, 20\%, \dots, 90\%$ where the last four cases of them are shown in (e)--(h). Fragmented trajectories are reconstructed using HDA with ReLu activation function, 1000 epochs, and the error tolerance of $10^{-6}$. Then, the same fragmented trajectories are reconstructed using LMC with 1000 iterations and the error tolerance of $10^{-6}$. (d) RMSE between the reconstructed and the original trajectories shows that both methods perform equally well when $p\le70$; but, HDA outperforms LMC when $p>70\%$. For the fragmentations $p=60\%$--$90\%$, (i)--(l) show the corresponding reconstructions by HDA and (m)--(p) show the corresponding reconstructions by LMC.}
\label{fig4}
\end{figure}

We approximate the linear rank as well as the nonlinear rank of this spiral dataset as explained in Sec.~\ref{sec:ranks}, where Fig.~\ref{fig4}(b) presents the percentages of SVs (computed using Eqn.~\eqref{eqn:per_SV}) for the linear approximation of the rank while Fig.~\ref{fig4}(c) presents the percentages of SVs for the nonlinear approximation of the rank. First, we observe that while the linear rank of the dataset is two, nonlinear rank of it is one. The nonlinear technique HDA should outperforms the linear technique LMC in reconstruction of trajectories since this dataset is naturally nonlinear. Second, we observe that $\tilde{\sigma}_2$ of this dataset is close to zero than $\tilde{\sigma}_2$ of the spiral dataset; thus, the nonlinear rank of this dataset is close to the linear rank in contrast to the spiral dataset that had a nonlinear rank which is clearly different from its linear rank. This leads better LMC's performance on this dataset than that of the spiral dataset. To empirically analyze these two conjectures, we use both LMC and HDA for the reconstruction of the fragmented trajectories. 

Similar to the spiral example, we generate fragmented trajectories by deleting variable percentages of the time-steps such that $p=10\%,\dots,90\%$, see Figs.~\ref{fig4}(e--h) for the last four fragmentations. To mimic real-life occlusion, we delete both of the coordinates of an agent if that time-step is to be deleted. We produce the binary indicator matrix, $\Omega$, to reflect the deleted time-steps of the data. Similar to the spiral dataset, we set the maximum epochs ($\eta$) to 1000, the error tolerance ($\tau$) to $10^{-6}$, and the activation function to ReLU, and train the HDA in Alg.~\ref{algo:HDA}(a) to get the trained weights, $\left\{W_k | \ k=1, \dots, K\right\}$, of the ANN. Alg.~\ref{algo:HDA}(b) receives the trained weights, $\left\{W_k | \ k=1, \dots, K\right\}$, as well as the indicator matrix and returns the reconstructed trajectories.

These fragmented trajectories are reconstructed again using the LMC technique presented in Alg.~\ref{algo:MC} with 1000 maximum iterations and $10^{-6}$ of error tolerance. We compute RMSE between the original trajectories and the reconstructed trajectories using Eqn.~\eqref{eq:rmse} that we show in Fig.~\ref{fig4}(d). Even though, both of the methods perform equally well when $p\le70\%$, HAD outperforms LMC when $p>70\%$. For the fragmentations $p=60\%$--$90\%$, we present the reconstructed trajectories of HDA in Figs.~\ref{fig4}(i--l) and of LMC in Figs.~\ref{fig4}(m--p). These figures justify that HDA performs well until $80\%$ of fragmentation while LMC only performs well until $70\%$ of fragmentation. 

We observed that the performance of HDA is better than that of LMC due to the fact that both the dataset and the HDA are nonlinear whereas LMC is linear. Moreover, we observed that the reconstruction performance of LMC in this dataset is better than that of the spiral dataset, since the nonlinear rank of this dataset is close to the linear rank in contrast to the nonlinear rank of the spiral dataset that is far to its linear rank.

\subsection{Mingling robots}\label{sec:robot}
\begin{figure}[htp]
\centering
\includegraphics[width=1\textwidth]{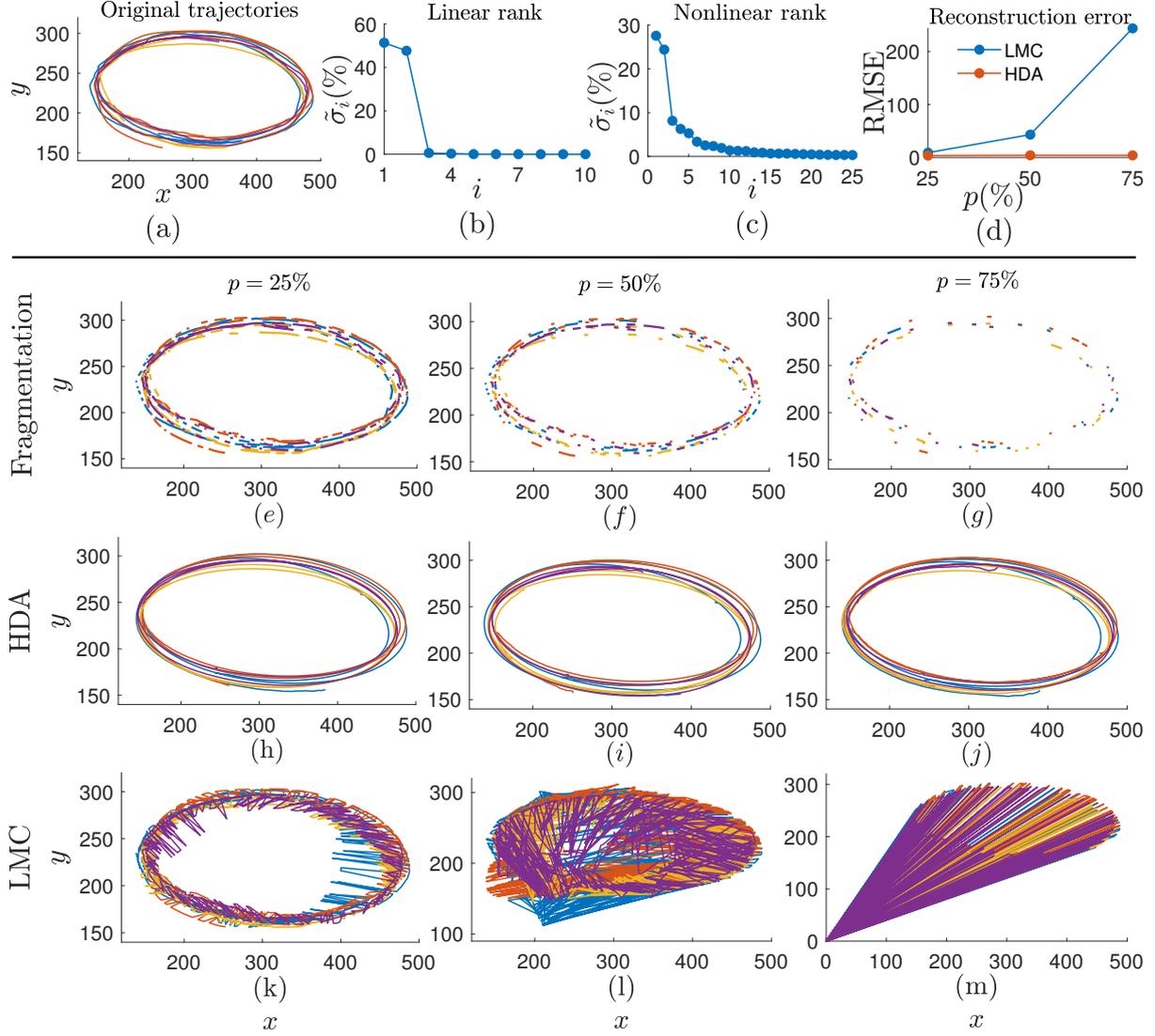}
\caption{Reconstruction of fragmented trajectories of four mingling robots. A video of mingling robots is tracked using a famous multi-object tracking method, Kanade–Lucas–Tomasi (KLT), and the individual trajectories of the agents are recorded, see (a). The percentages of the singular values (SVs) of, (b) a linear surrogate to rank of the dataset implies two whereas that of, (c) a nonlinear surrogate to rank implies a value around 20. We generate three datasets by deleting variable percentages, say $p\%$'s, of the trajectories such that $p=25\%$, $50\%$, and $75\%$ that we show in (e)--(g). The fragmented trajectories are reconstructed using HDA with ReLu as the activation function, 1000 epochs, and the error tolerance of $10^{-6}$. Then, we reconstruct the same fragmented trajectories using LMC with 1000 iterations and the error tolerance of $10^{-6}$. (d) RMSE between the reconstructed and the original trajectories justifies that HDA performs exceptionally better than LMC for all the fragmentation levels. While (h)--(j) show the trajectories reconstructed by HDA, (k)--(m) show the corresponding trajectories reconstructed by LMC.}
\label{fig5}
\end{figure}

As the last example, we consider a real-life collective motion scenario where four robots mingle in a circular pattern that we downloaded from \cite{Benedettelli2009}. We track the trajectories of the robots in this video, see Fig.~\ref{fig5}(a), using a famous MOT method, called KLT tracker, available in \cite{lucas1981iterative, tomasi1991detection}, with its default parameter values. We approximate both the linear rank and the nonlinear rank of the this dataset as explained in Sec.~\ref{sec:ranks}, where the percentages of SVs for the linear and the nonlinear approximations are presented in Fig.~\ref{fig5}(b) and Fig.~\ref{fig5}(c), respectively. Here, we observe that while the rank associated with the linear approximation of the dataset is two, that with nonlinear approximation is somewhere around 20. Nonlinearity of this dataset is far away from its linearity due to the factors such as less coordination of agents' motion and less precision of the KLT tracker. We can conjecture that this significant difference between two types of ranks will grant drastically different reconstruction performance between the linear methods LMC and the nonlinear methods HDA. To empirically analyze this conjecture, we use both LMC and HDA for the reconstruction of the robot data.

Similar to the two previous examples, we generate fragmented trajectories by deleting variable percentages, say $p$'s, of the time-steps of the trajectories such that $p=25\%$, $50\%$, and $75\%$, see Figs.~\ref{fig5}(e--h). The binary indicator matrix, $\Omega$, that we create here reflects the deleted time-steps of the trajectory data by zeros. Similar to the two previous examples, we set the maximum epochs ($\eta$) to 1000, the error tolerance ($\tau$) to $10^{-6}$, and the activation function to ReLU, and train the HDA in Alg.~\ref{algo:HDA}(a). The trained weights, $\left\{W_k | \ k=1, \dots, K\right\}$, along with the indicator matrix is fed into Alg.~\ref{algo:HDA}(b) to obtain the reconstructed trajectories. Alg.~\ref{algo:MC} of LMC is implemented with 1000 of maximum iterations and $10^{-6}$ of error tolerance to reconstruct fragmented trajectories. RMSE is computed between the original trajectories and the reconstructed trajectories using Eqn.~\eqref{eq:rmse} that we shown in Fig.~\ref{fig3}(d). RMSEs justify the initial conjecture that HDA should significantly outperforms LMC. This example provides a salient empirical result in the support of the validity of HDA over LMC in reconstruction of nonlinear real-life trajectory data.

\section{Discussion}\label{sec:diss}
MOT plays an important role in collective motion studies since trajectories facilitate the learning of the dynamics of not only the entire group, but also that of either each individual agent or their clusters \cite{Gajamannage2015}. The trajectories constructed by the conventional MOT methods might lose some segments of the trajectories due to natural phenomenon such as occlusion, change of illumination, etc. \cite{Jia12}. In this paper, we have presented a machine learning approach, named as Hadamard deep autoencoder (HDA), for the reconstruction of fragmented trajectories of collective motion and then compared the performance of this method with another machine learning technique, called low-rank matrix completion (LMC), for trajectory reconstruction that we presented in \cite{Gajamannage2019Reconstruction}. Each agent practicing collective motion, is interacted with its neighbors within a limited radius, grants interesting properties to its trajectory dataset including low-rankness. We utilized this low-rank property to reconstruct fragmented trajectories and empirically investigated the influence of the datasets' nonlinearity for the performance of HDA and LMC. We simulated two synthetic collective motion datasets using a self-propelled swarming model, called the Vicsek model \cite{Vicsek1995}, and use a real-life collective motion video as the third dataset.

Conventional MOT methods treat the fragmentation issue of the tracked trajectories using inbuilt, but computationally expensive, trajectory defragmentation methods such as \emph{part-to-whole}, \emph{hypothesize-and-test}, and \emph{ buffer-and-recover}\cite{wenhan2021}. We are unable to compare/contrast our trajectory defragmentation technique with existing defragmentation techniques as they are not available separately out of the tracking methods. We are planning to improve the MOT method presented as a part of the study presented in \cite{Gajamannage2015} by prepending it onto our trajectory defragmentation technique to produce a complete MOT method. This future work allows us to compare our complete MOT method with the existing MOT methods such as the ones presented in \cite{lucas1981iterative,tomasi1991detection, Fontaine07,yang11,koller12,lu13}. Especially, the simple MOT method presented in \cite{Crocker99} was used to track colloidal particles efficiently. We have observed that this method is capable of tracking trajectories of agents in collective motion with some  fragmentations at some time-steps of the trajectories. Thus, appending our trajectory defragmentation method onto this colloidal particles tracker originates an efficient MOT method for the literature of computer vision. Such MOT methods that are especially focused on agent tracking in collective motion will be helpful for researchers to study the dynamics of diverse types of real-life collective motion scenarios recorded as video.

The performance analysis in this paper covers three examples where two of them are based on collective motion simulations, a particle swarm imitating a spiral-shaped collective motion and another particle swarm imitating collective motion at an obstacle, and the other is based on a real-life collective motion scenario of a mingling robots swarm. While we observed in the first two examples that HDA and LMC are competitive until with $70\%$ of fragmentation, HDA outperforms LMC with more fragmentation. However, we observed that while HDA performs exceptionally well for the robot dataset, the performance of LMC was significantly weak. HDA is a nonlinear technique for trajectory reconstruction whereas LMC is a linear technique for the same task. We observed that  the linear approximations of the ranks are two, two with a small second SV, and two for the datasets spiral, obstacle, and robot, respectively, whereas the nonlinear approximations of the ranks are one, one, and around 20 for the datasets spiral, obstacle, and robot, respectively. The order from the smallest to the largest of the difference between the linear approximation of the rank and the nonlinear approximation of the rank of the datasets is obstacle, spiral, and robots. Moreover, the order from the best to the worst of the performance of HDA over LMC is for the datasets obstacle, spiral, and robots. Since these two orders are the same, we can conclude that more nonlinear the trajectory data better the performance of HDA over LMC. Collective motion is non-linear in general; thus, HDA is a vital technique for the reconstruction of fragmented trajectories of collective motion data. 

As we have seen in Sec.~\ref{sec:perAna}, while linear and nonlinear approximations of rank of the simulated trajectory datasets are close to each other, that of real-life trajectory datasets are far apart. This improves the performance of HDA exceptionally for real-life collective motion datasets in contrast to that of LMC. Real-life trajectory data attains big non-linear ranks since the agents in natural collective motion scenarios are not perfectly coordinated like the agents in collective motion simulations. Thus, it is always better to implement HDA instead of LMC if the trajectory data of interest is from real-life collective motion instances. 

We presented a nonlinear machine learning technique for fragmented trajectory reconstruction in this paper which is made by incorporating an indicator matrix into a regular deep autoencoder using Hadamard product. This binary indicator matrix has ones for the observed time-steps of the trajectories and zeros elsewhere. Trajectories of the agents practicing collective motion is nonlinear; thus, nonlinear nature of HDA is the vital property for its performance in contrast to linear methods such as LMC. We empirically justified that our HDA possesses superior performance even for $80\%$ of the trajectory fragmentation. While LMC is unable of producing faithful reconstructions for fragmented real-life trajectories, HDA remains attaining its superior reconstruction performance for such trajectories. 

\section{Acknowledgments}
The authors would like to thank the Google Cloud Platform for granting Research Credit to access its GPU computing resources under the project number 397744870419.


\end{document}